\documentclass[10pt,twocolumn,letterpaper]{article}

 \usepackage{cvpr}              % To produce the CAMERA-READY version
\usepackage{multirow}
\usepackage{algorithm}
\usepackage{algpseudocode}
\DeclareMathOperator*{\argmin}{arg\,min}

\usepackage[dvipsnames]{xcolor}

\definecolor{cvprblue}{rgb}{0.21,0.49,0.74}
\usepackage[pagebackref,breaklinks,colorlinks,citecolor=cvprblue]{hyperref}

\def\paperID{00011} % *** Enter the Paper ID here
\def\confName{CVPR}
\usepackage{bm}
\def\confYear{2024}

\title{ATOM: Attention Mixer for Efficient Dataset Distillation}

\author{Samir Khaki$^{1}$\thanks{Equal contribution}, ~Ahmad Sajedi$^{1*}$, ~Kai Wang$^{2}$, ~Lucy Z. Liu$^{3}$, ~Yuri A. Lawryshyn$^{1}$,\\ and ~Konstantinos N. Plataniotis$^{1}$\\
	 $^{1}$University of Toronto ~~~~~~~ $^{2}$ National University of Singapore ~~~~~~~ $^{3}$Royal Bank of Canada (RBC)\\
	{\tt\small \{samir.khaki, ahmad.sajedi\}@mail.utoronto.ca}\\
 \tt\small Code: \href{https://github.com/DataDistillation/ATOM}{https://github.com/DataDistillation/ATOM}} 
\begin{document}
\maketitle
\begin{abstract}
Recent works in dataset distillation seek to minimize training expenses by generating a condensed synthetic dataset that encapsulates the information present in a larger real dataset. These approaches ultimately aim to attain test accuracy levels akin to those achieved by models trained on the entirety of the original dataset. Previous studies in feature and distribution matching have achieved significant results without incurring the costs of bi-level optimization in the distillation process. Despite their convincing efficiency, many of these methods suffer from marginal downstream performance improvements, limited distillation of contextual information, and subpar cross-architecture generalization. To address these challenges in dataset distillation, we propose the \textbf{AT}tenti\textbf{O}n \textbf{M}ixer (\textbf{ATOM}) module to efficiently distill large datasets using a mixture of channel and spatial-wise attention in the feature matching process. Spatial-wise attention helps guide the learning process based on consistent localization of classes in their respective images, allowing for distillation from a broader receptive field. Meanwhile, channel-wise attention captures the contextual information associated with the class itself, thus making the synthetic image more informative for training. By integrating both types of attention, our ATOM module demonstrates superior performance across various computer vision datasets, including CIFAR10/100 and TinyImagenet. Notably, our method significantly improves performance in scenarios with a low number of images per class, thereby enhancing its potential. Furthermore, we maintain the improvement on cross-architectures and applications such as neural architecture search.
\end{abstract}    
\section{Introduction}
\label{sec:intro}

\begin{figure}[t]
    \centering
    \includegraphics[width=0.45\textwidth]{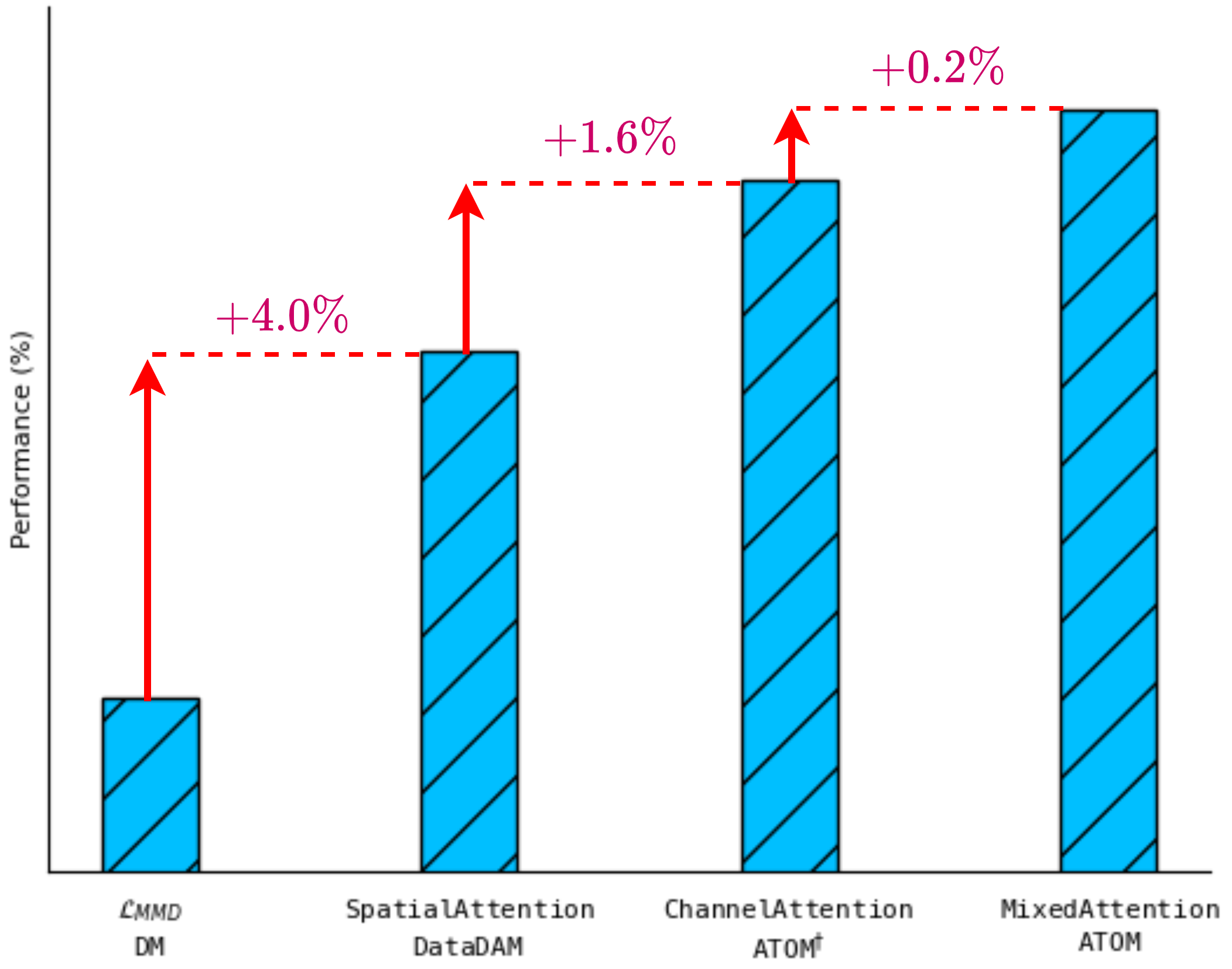}
    \caption{The \texttt{ATOM} Framework utilizes inherent information to capture both context and location, resulting in significantly improved performance in dataset distillation. We display the performance of various components within the  \texttt{ATOM} framework, showcasing a $5.8\%$ enhancement from the base distribution matching performance on CIFAR10 at IPC50. Complete numerical details can be found in \Cref{tab:component}.}
    \label{fig:teaser}
\end{figure}

Efficient deep learning has surged in recent years due to the increasing computational costs associated with training and inferencing pipelines \cite{hinton2015distilling, wang2018dataset, zhao2021dataset, sajedi2023new, zhao2023dataset, sajedi2023datadam, yu2017compressing, sajedi2023endtoend, sajedi2024probmcl, amer2021high}. This growth can be attributed to the escalating complexity of model architectures and the ever-expanding scale of datasets. Despite the increasing computational burden, two distinct approaches have emerged as potential avenues for addressing this issue: the model-centric and data-centric approaches. The model-centric approach is primarily concerned with mitigating computational costs by refining the architecture of deep learning models. Techniques such as pruning, quantization, knowledge distillation, and architectural simplification are key strategies employed within this paradigm \cite{hinton2015distilling, wu2016quantized, yu2017compressing, 9679989, khaki2023cfdp, khaki2024need, sajedi2022subclass, sajedi2021efficiency}. In contrast, the data-centric approach adopts a different perspective, focusing on exploring and leveraging the inherent redundancy within datasets. Rather than modifying model architectures, this approach seeks to identify or construct a smaller dataset that retains the essential information necessary for maintaining performance levels. Coreset selection was a fairly adopted method for addressing this gap \cite{rebuffi2017icarl, castro2018end, belouadah2020scail, sener2018active, tonevaempirical}. In particular works such as Herding \cite{welling2009herding} and K-Center \cite{sener2018active} offered a heuristic-based approach to intelligently select an informative subset of data. However, as a heuristic-based method, the downstream performance is limited by the information contained solely in the subset. More recently, shapely data selection \cite{ghorbani2022data} found the optimal subset of data by measuring the downstream performance for every subset combination achievable in the dataset. However inefficient this may be, the downstream performance is still limited by the diversity of samples selected. therefore, Dataset Distillation (DD) \cite{wang2018dataset} has emerged as a front-runner wherein a synthetic dataset can be learned.

Dataset distillation aims to distill large-scale datasets into a smaller representation, such that downstream models trained on this condensed dataset will retain competitive performance with those trained on the larger original one \cite{wang2018dataset, zhao2021dataset, cazenavette2022dataset}. Recently, many techniques have been introduced to address this challenge, including gradient matching \cite{zhao2021dataset, zhao2021dataset2, lee2022dataset}, feature/distribution matching \cite{zhao2023dataset, sajedi2023datadam, zhao2023improved}, and trajectory matching \cite{cazenavette2022dataset, du2023minimizing, guo2024towards}. However, many of these methods suffer from complex and computationally heavy distillation pipelines \cite{zhao2021dataset, cazenavette2022dataset, guo2024towards} or inferior performance \cite{zhao2023dataset, sajedi2023datadam, zhao2021dataset}. A promising approach, DataDAM \cite{sajedi2023datadam}, effectively tackled the computational challenges present in prior distillation techniques by employing untrained neural networks, in contrast to bi-level optimization methods. However, despite its potential, DataDAM faced several significant limitations: (1) it obscured relevant class-content-based information existing channel-wise in intermediate layers; (2) it only achieved marginal enhancements on previous dataset distillation algorithms; and (3) it exhibited inferior cross-architecture generalization.

% A promising work. DataDAM \cite{sajedi2023datadam} efficiently addressed the computational bottlenecks in previous distillation methods by using untrained neural networks, similar to \cite{} as opposed to bi-level optimization methods. Despite its potential, there were a few major limitations, including: (1) obscuring the relevant class-content-based information that exists channel-wise in the intermediate layers, (2) marginal improvements on the previous distillation test suite, and (3) inferior cross-architecture generalization.

In this work, we introduce \textbf{AT}tenti\textbf{O}n \textbf{M}ixer, dubbed \texttt{ATOM} as an efficient dataset distillation pipeline that strikes an impressive balance between computational efficiency and superior performance. Drawing upon spatial attention matching techniques from prior studies like DataDAM \cite{sajedi2023datadam}, we expand our receptive field of information in the matching process. Our key contribution lies in mixing spatial information with channel-wise contextual information. Intuitively, different convolutional filters focus on different localizations of the input feature; thus, channel-wise attention aids in the distillation matching process by compressing and aggregating information from multiple regions as evident by the performance improvmenets displayed in \Cref{fig:teaser}. \texttt{ATOM} not only combines localization and context, but it also produces distilled images that are more generalizable to various downstream architectures, implying that the distilled features are true representations of the original dataset. Moreover, our approach demonstrates consistent improvements across all settings on a comprehensive distillation test suite. In summary, the key contributions of this study can be outlined as follows:
% In this work, we introduce \texttt{ATOM} as an efficient distillation pipeline that achieved a stellar trade-off between computational complexity and superior performance. We leverage the spatial attention matching from previous works such as DataDAM \cite{} to increase our receptive field of information in the matching process, however, our key contribution is mixing spatial information with channel-wise contextual information. Intuitively, different convolutional filters may capture/focus on different localization in the input feature, hence channel-wise attention helps to compress and aggregate information from different regions, thus aiding the distillation matching process. ATOM not only mixes localization with context but generates distilled data that is far more generalizable to various downstream architectures, indicating that the distilled features are truly representative of the base dataset. Further, we improve in all categories on the larger distillation test suite.  The contributions of this study are summarized as:

\textbf{[C1]}: We provide further insight into the intricacies of attention matching, ultimately introducing the use of channel-wise attention matching for capturing a higher level of information in the feature-matching process. Our mixing module combines both spatial localization awareness of a particular class, with distinctive contextual information derived channel-wise.
% We provide further insight into the intricacies of attention matching, ultimately introducing the use of channel-wise attention matching for capturing a greater degree of information in the feature-matching process. Our mixing module combines both spatial localization awareness of a particular class, with distinctive contextual information derived channel-wise.

\textbf{[C2]}: Empirically we show superior performance against previous dataset distillation methods including feature matching and attention matching works, without bi-level optimization on common computer vision datasets.

\textbf{[C3]}: We extend our findings by demonstrating superior performance in cross-architecture and neural architecture search. In particular, we provide a channel-only setting that maintains the majority of the performance while incurring a lower computational cost.
% We extend our results with superior performance in cross-architecture, and neural architecture search, all with a better tradeoff in GPU memory consumption as opposed to previous attention and feature matching works. In particular, we offer a channel-only setting that retains the majority of the performance with a lower computational cost.

\section{Related Works}
\label{sec:related}

\textbf{Coreset Selection.} Coreset selection, an early data-centric approach, aimed to efficiently choose a representative subset from a full dataset to enhance downstream training performance and efficiency. Various methods have been proposed in the past, including geometry-based approaches \cite{agarwal2020contextual, chen2010super, sener2018active, sinha2020small, welling2009herding}, loss-based techniques as mentioned in \cite{toneva2018empirical, paul2021deep}, decision-boundary-focused methods \cite{margatina2021active, ducoffe2018adversarial}, bilevel optimization strategies \cite{killamsetty2021glister, killamsetty2021retrieve}, and gradient-matching algorithms outlined in \cite{mirzasoleiman2020coresets, killamsetty2021grad}. Notable among them are Random, which randomly selects samples as the coreset; Herding, which picks samples closest to the cluster center; K-Center, which selects multiple center points to minimize the maximum distance between data points and their nearest center; and Forgetting, which identifies informative training samples based on learning difficulties \cite{castro2018end, belouadah2020scail, sener2018active, toneva2018empirical}. While these selection-based methods have shown moderate success in efficient training, they inherently possess limitations in capturing rich information. Since each image in the selected subset is treated independently, they lack the rich features that could have been captured if the diversity within classes had been considered. These limitations have motivated the emergence of dataset distillation within the field.
%Coreset selection was an early introduced data-centric approach that intelligently selects a representative subset from the full dataset such that downstream training can achieve competitive performance. Previous methods have proposed various heuristic selection criteria, such as geometry-based \cite{agarwal2020contextual, chen2010super, sener2018active, sinha2020small, welling2009herding}, loss-based \cite{toneva2018empirical, paul2021deep}, decision-boundary-based \cite{margatina2021active, ducoffe2018adversarial}, bilevel optimization \cite{killamsetty2021glister, killamsetty2021retrieve}, and gradient-matching \cite{mirzasoleiman2020coresets, killamsetty2021grad}. Among them, Random selects samples randomly as the coreset; Herding \cite{castro2018end, belouadah2020scail} picks the closest samples to the cluster center; K-Center \cite{sener2018active} chooses multiple center points to minimize the maximum distance between a data point and its nearest center and Forgetting \cite{toneva2018empirical} identifies informative training samples based on their difficulties in learning. Despite moderate success in outperforming the random selection baseline, selection-based methods are inherently limited in the information they can capture. Since each image in the selected subset is independent, they lack information-rich features that could have been captured had the diversity of the class been considered. Ultimately the limitations of this method inspired the field dataset distillation.
\begin{figure*}[t]
\centering
    {\includegraphics[width=1\textwidth]{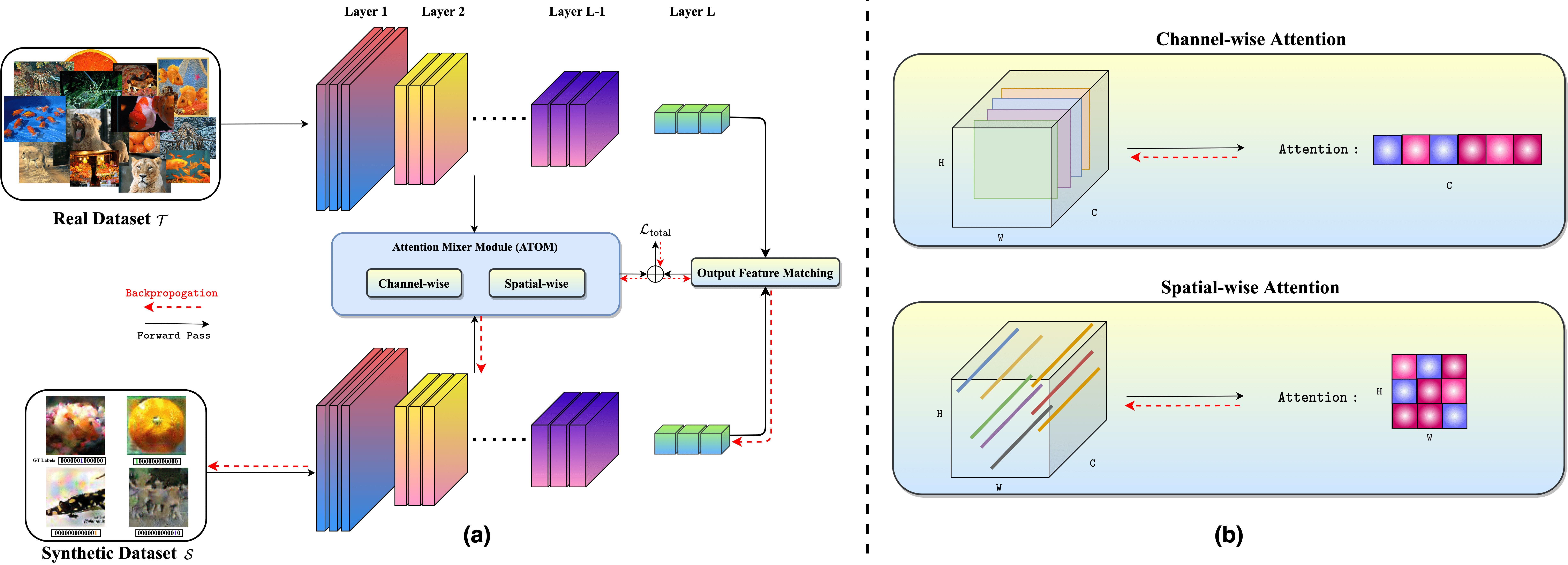}}
    \caption{(a) An overview of the proposed \texttt{ATOM} framework. By mixing attention, \texttt{ATOM} is able to capture both spatial localization and class context. (b) Demonstration of the internal architecture for spatial- and channel-wise attention in the \texttt{ATOM} Module. The spatial-wise attention computes attention at specific locales through different filters, resulting in a matrix output, whereas the channel-wise attention calculates attention between each filter, naturally producing a vectorized output.}
    \label{fig:ATOM}
\end{figure*}

\textbf{Dataset Distillation.} Dataset distillation has emerged as a learnable method of synthesizing a smaller, information-rich dataset from a large-scale real dataset. This approach offers a more efficient training paradigm, commonly applied in various downstream applications such as continual learning \cite{chen2024data, sajedi2023datadam, zhao2021dataset, gu2024ssd, yang2024efficient}, neural architecture search \cite{ho2016generative, such2020generative}, and federated learning \cite{jia2023unlocking, xiong2023feddm, liu2023meta, liu2023slimmable}. The seminal work, initially proposed by Wang et al. \cite{wang2018dataset}, introduced bilevel optimization, comprising an outer loop for learning the pixel-level synthetic dataset and an inner loop for training the matching network. Following this, several studies adopted surrogate objectives to tackle unrolled optimization problems in meta-learning. For example, gradient matching methods \cite{zhao2021dataset, zhao2021dataset2, lee2022dataset, kim2022dataset, du2024sequential} learn images by aligning network gradients derived from real and synthetic datasets. Trajectory matching \cite{cazenavette2022dataset, du2023minimizing, cui2023scaling, guo2024towards} improves performance by minimizing differences in model training trajectories between original and synthetic samples. Meanwhile, feature matching strategies \cite{zhao2023dataset, wang2022cafe, sajedi2023datadam, zhao2023improved, zhang2024m3d, sajedi2023datadam} aim to align feature distributions between real and synthetic data within diverse latent spaces. Despite significant advancements in this field, methods still struggle to find a trade-off between the computational costs associated with the distillation pipeline and the model's performance. A recent work, DataDAM \cite{sajedi2023datadam}, used spatial attention to improve the performance of feature-matching-based methods by selectively matching features based on their spatial attention scores. However, although this method operates without bilevel optimization, it only marginally improves performance on larger test suites. In this study, we delve deeper into the potential of attention-based methods and demonstrate superior performance compared to DataDAM and previous benchmarks across various computer vision datasets. Additionally, we achieve a lower computational cost compared to conventional attention-matching approaches by leveraging information in a channel-wise manner.

\textbf{Attention Mechanism.} Attention mechanisms have been widely adopted in deep learning to enhance performance across various tasks \cite{bahdanau2015neural, wang2018non, zagoruyko2016paying}. Initially applied in natural language processing \cite{bahdanau2015neural}, it has extended to computer vision, with global attention models \cite{wang2018non} improving image classification and convolutional block attention modules \cite{woo2018cbam} enhancing feature map selection. Additionally, attention aids model compression in knowledge distillation \cite{zagoruyko2016paying}. They are lauded for their ability to efficiently incorporate global contextual information into feature representations. When applied to feature maps, attention can take the form of either spatial or channel-based methods. Spatial methods focus on identifying the informative regions ("where"), while channel-based methods complementarily emphasize the informative features ("what"). Both spatial localization and channel information are crucial for identifying class characteristics. Recently, Sajedi \etal proposed DataDAM \cite{sajedi2023datadam} to concentrate only on spatial attention, capturing class correlations within image localities for efficient training purposes. However, inspired by the inherent obfuscation of the content in the attention maps, we propose an Attention Mixer module that uses a unique combination of spatial and channel-wise attention to capture localization and information content.
\section{Methodology}
\label{sec:method}

Given the larger source dataset $\mathcal{T} = \{(\bm{x}_{i}, y_{i})\}_{i=1}^{|\mathcal{T}|}$ containing $|\mathcal{T}|$ real image-label pairs, we generate a smaller learnable synthetic dataset $\mathcal{S} = \{(\bm{s}_{j}, y_{j})\}_{j=1}^{|\mathcal{S}|}$ with $|\mathcal{S}|$ synthetic image and label pairs. Following previous works \cite{zhao2021dataset, zhao2021dataset2, wang2022cafe, sajedi2023datadam, cazenavette2022dataset}, we use random sampling to initialize our synthetic dataset. 
For every class $k$, we obtain a batch of real and synthetic data ($B^{\mathcal{T}}_{k}$ and $B^{\mathcal{S}}_{k}$, respectively) and use a neural network $\phi_{\boldsymbol{\theta}}(\cdot)$ with randomly initialized weights $\boldsymbol{\theta}$ \cite{he2015delving} to extract intermediate and output features. We illustrate our method in \Cref{fig:ATOM} where an $L$-layer neural network $\phi_{\boldsymbol{\theta}}(\cdot)$ is used to extract features from the real and synthetic sets. The collection of feature maps from the real and synthetic sets can be expressed as $\phi_{\boldsymbol{\theta}}({\mathcal{T}}_{k}) = [\bm{f}^{\mathcal{T}_{k}}_{\boldsymbol{\theta},1}, \cdots, \bm{f}^{\mathcal{T}_{k}}_{\boldsymbol{\theta}, L}]$ and $\phi_{\boldsymbol{\theta}}(\mathcal{S}_{k}) = [\bm{f}^{\mathcal{S}_{k}}_{\boldsymbol{\theta},1}, \cdots, \bm{f}^{\mathcal{S}_{k}}_{\boldsymbol{\theta}, L}]$, respectively. The feature $\bm{f}^{\mathcal{T}_{k}}_{\boldsymbol{\theta},l}$ comprises a multi-dimensional array within $\mathbb{R}^{|B^{\mathcal{T}}_{k}| \times C_{l} \times W_{l}\times H_{l}}$, obtained from the real dataset at the $l^\text{th}$ layer, where $C_{l}$ denotes the number of channels and $H_{l} \times W_{l}$ represents the spatial dimensions. Correspondingly, a feature $\bm{f}^{\mathcal{S}_{k}}_{\boldsymbol{\theta},l}$ is derived for the synthetic dataset.

We now introduce the Attention Mixer Module (\texttt{ATOM}) which generates attention maps for the intermediate features derived from both the real and synthetic datasets. Leveraging a feature-based mapping function $A(\cdot)$, \texttt{ATOM} takes the intermediate feature maps as input and produces a corresponding attention map for each feature. Formally, we express this as: $A\big(\phi_{\boldsymbol{\theta}}({\mathcal{T}}_{k})\big) = [\bm{a}^{\mathcal{T}_{k}}_{\boldsymbol{\theta},1}, \cdots, \bm{a}^{\mathcal{T}_{k}}_{\boldsymbol{\theta}, L-1}]$ and $A(\phi_{\boldsymbol{\theta}}({\mathcal{S}}_{k})) = [\bm{a}^{\mathcal{S}_{k}}_{\boldsymbol{\theta},1}, \cdots, \bm{a}^{\mathcal{S}_{k}}_{\boldsymbol{\theta}, L-1}]$ for the real and synthetic sets, respectively. Previous works \cite{sajedi2023datadam, zagoruyko2016paying} have shown that spatial attention, which aggregates the absolute values of feature maps across the channel dimension, can emphasize common spatial locations associated with high neuron activation. The implication of this is retaining the most informative regions, thus generating an efficient feature descriptor. In this work, we also consider the effect of channel-wise attention, which emphasizes the most significant information captured by each channel based on the magnitude of its activation. Since different filters explore different regions or locations of the input feature, channel-wise activation yields the best aggregation of the global information. Ultimately, we convert the feature map $\bm{f}^{\mathcal{T}_{k}}_{\boldsymbol{\theta},l}$ of the $l^\text{th}$ layer into an attention map $\bm{a}^{\mathcal{T}_{k}}_{\boldsymbol{\theta},l}$ representing spatial or channel-wise attention using the corresponding mapping functions $A_{s}(\cdot)$ or $A_{c}(\cdot)$ respectively. Formally, we can denote the spatial and channel-wise attention maps as:

\begin{flalign} \label{eq:spatial}
 A_s(\bm{f}^{\mathcal{T}_{k}}_{\boldsymbol{\theta},l}) = \sum_{i=1}^{C_{l}}\big|{(\bm{f}^{\mathcal{T}_{k}}_{\boldsymbol{\theta},l})}_{i}\big|^{p_{s}},
\end{flalign}
\begin{flalign} \label{eq:channel}
 A_c(\bm{f}^{\mathcal{T}_{k}}_{\boldsymbol{\theta},l}) = \sum_{i=1}^{H_{l}*W_{l}}\big|(\bm{f}^{\mathcal{T}_{k}}_{\boldsymbol{\theta},l})^\star_{i}\big|^{p_{c}},
\end{flalign}
where, ${(\bm{f}^{\mathcal{T}_{k}}_{\boldsymbol{\theta},l})}_{i} = \bm{f}^{\mathcal{T}_{k}}_{\boldsymbol{\theta},l} (:,i,:,:) $ is the feature map of channel $i$ from the $l^\text{th}$ layer, and the power and absolute value operations are applied element-wise; meanwhile, the symbol $^\star$ flattens the feature map along the spatial dimension $\left((\bm{f}^{\mathcal{T}_{k}}_{\boldsymbol{\theta},l})^* \in \mathbb{R}^{|B^{\mathcal{T}}_{k}| \times C_{l} \times W_{l}*H_{l}} \right)$, such that $(\bm{f}^{\mathcal{T}_{k}}_{\boldsymbol{\theta},l})^\star_{i} = (\bm{f}^{\mathcal{T}_{k}}_{\boldsymbol{\theta},l})^{\star} (:,:,i)$. By leveraging both types of attention, we can better encapsulate the relevant information in the intermediate features, as investigated in \Cref{sec:ablate_loss_components}. Further, the effect of power parameters for spatial and channel-wise attention, \ie $p_{s}$ and $p_{c}$ is studied in the \Cref{sec:ablation_power}.

Given our generated spatial and channel attention maps for the intermediate features, we apply standard normalization such that we can formulate a matching loss between the synthetic and real datasets. We denote our generalized loss $\mathcal{L}_\text{ATOM}$ as: 
\begin{flalign} \label{eq:atom}
\displaystyle \mathop{\mathbb{E}}_{\boldsymbol{\theta}\sim P_{\boldsymbol{\theta}}}\bigg[\sum_{k=1}^{K}\sum_{l=1}^{L-1}\Big\lVert \displaystyle {\mathbb{E}}_{\mathcal{T}_{k}}\Big[\frac{\bm{z}^{\mathcal{T}_{k}}_{\boldsymbol{\theta}, l}}{{\lVert \bm{z}^{\mathcal{T}_{k}}_{\boldsymbol{\theta}, l}\rVert}_{2}}\Big] - \displaystyle \mathbb{E}_{{\mathcal{S}}_{k}}\Big[\frac{\bm{z}^{\mathcal{S}_{k}}_{\boldsymbol{\theta}, l}}{{\lVert \bm{z}^{\mathcal{S}_{k}}_{\boldsymbol{\theta}, l}\rVert}_{2}}\Big]\Big\rVert^{2}\bigg],
\end{flalign}
where, in the case of spatial attention, we denote $\bm{z}^{\mathcal{T}_{k}}_{\boldsymbol{\theta}, l} = vec(\bm{a}^{\mathcal{T}_{k}}_{\boldsymbol{\theta},l}) \in \mathbb{R}^{|B^{\mathcal{T}}_{k}| \times (W_{l}\times H_{l})}$ and $\bm{z}^{\mathcal{S}_{k}}_{\boldsymbol{\theta}, l} = vec(\bm{a}^{\mathcal{S}_{k}}_{\boldsymbol{\theta},l}) \in \mathbb{R}^{|B^{\mathcal{S}}_{k}|\times (W_{l}\times H_{l})}$ to represent the vectorized spatial attention map pairs at the $l^\text{th}$ layer for the real and synthetic datasets, respectively. Meanwhile, for channel-based attention, we have
$\bm{z}^{\mathcal{T}_{k}}_{\boldsymbol{\theta}, l} = vec(\bm{a}^{\mathcal{T}_{k}}_{\boldsymbol{\theta},l}) \in \mathbb{R}^{|B^{\mathcal{T}}_{k}| \times (C_{l})}$ and $\bm{z}^{\mathcal{S}_{k}}_{\boldsymbol{\theta}, l} = vec(\bm{a}^{\mathcal{S}_{k}}_{\boldsymbol{\theta},l}) \in \mathbb{R}^{|B^{\mathcal{S}}_{k}|\times (C_{l})}$ to represent the flattened channel attention map pairs at the $l^\text{th}$ layer for the real and synthetic datasets, respectively. The parameter $K$ is the number of categories in a dataset, and $P_{\boldsymbol{\theta}}$ denotes the distribution of network parameters. We estimate the expectation terms in \Cref{eq:atom} empirically if ground-truth data distributions are not available.

Following previous works \cite{sajedi2023datadam, zhao2023dataset, wang2022cafe, zhao2023improved, zhang2024m3d}, we leverage the features in the final layer to regularize our matching process. In particular, the features of the penultimate layer represent a high-level abstraction of information from the input images in an embedded representation and can thus be used to inject semantic information in the matching process \cite{sajedi2023datadam, saito2018maximum, zhao2023dataset, gretton2012kernel}. Thus, we employ $\mathcal{L}_{\text{MMD}}$ as described in \cite{sajedi2023datadam, zhao2023dataset} out-of-the-box. 

Finally, we learn the synthetic dataset by minimizing the following optimization problem using SGD optimizer:
\begin{flalign} \label{eq:loss}
\mathcal{S}^{*} = \argmin_{\mathcal{S}}\:\big(\mathcal{L}_{\text{ATOM}} + \lambda \mathcal{L}_{\text{MMD}}\big),
\end{flalign}
where $\lambda$ is the task balance parameter inherited from \cite{sajedi2023datadam}. In particular, we highlight that $\mathcal{L}_\text{MMD}$ brings semantic information from the final layer, while $\mathcal{L}_\text{ATOM}$ mixes the spatial and channel-wise attention information from the intermediate layers. Note that our approach assigns a fixed label to each synthetic sample and keeps it constant during training. A summary of the learning algorithm can be found in Algorithm \ref{alg:1}.

\begin{algorithm}[h] 
\caption{Attention Mixer for Dataset Distillation}
\label{alg:1}
\textbf{Input:} \text{Real training dataset $\mathcal{T}=\{(\bm{x}_{i}, y_{i})\}_{i=1}^{|\mathcal{T}|}$}\\
\textbf{Required:} Initialized synthetic samples for $K$ classes, Deep neural network $\phi_{\bm{\theta}}$ parameterized with $\boldsymbol{\bm{\theta}}$, Probability distribution over randomly initialized weights $P_{\boldsymbol{\theta}}$, Learning rate $\eta_{\mathcal{S}}$, Task balance parameter $\lambda$, Number of training iterations $I$.%, where $|\mathcal{S}| \ll |\mathcal{T}|$
\begin{algorithmic}[1]
\State Initialize synthetic dataset $\mathcal{S}$ %\algorithmiccomment{\textcolor{blue}{Use Gaussian noise, Random sampling or K-Center\cite{cuidc} from a real dataset}}
\For{$i = 1, 2, \cdots, I$}  %\label{line:1}
	   \State Sample $\bm{\theta}$ from $P_{\bm{\theta}}$
        \State Sample mini-batch pairs $B_{k}^{\mathcal{T}}$ and $B_{k}^{\mathcal{S}}$ from the real 
        \Statex \:\:\:\:\:\: and synthetic sets for each class $k$
        \State Compute $\mathcal{L}_{\text{ATOM}}$ and $\mathcal{L}_{\text{MMD}}$ 
        \State Calculate $\mathcal{L} = \mathcal{L}_{\text{ATOM}} + \lambda \mathcal{L}_{\text{MMD}}$
        \State Update the synthetic dataset using $\mathcal{S} \leftarrow \mathcal{S} - \eta_{\mathcal{S}}\nabla_{\mathcal{S}}\mathcal{L}$
\EndFor
\end{algorithmic}

\textbf{Output:} \text{Synthetic dataset $\mathcal{S}=\{(\bm{s}_{i}, y_{i})\}_{i=1}^{|\mathcal{S}|}$}

\end{algorithm}
\section{Experiments} \label{sec:exp}

\subsection{Experimental Setup} \label{exp-set}

\textbf{Datasets.} Our method is evaluated on the CIFAR-10 and CIFAR-100 datasets \cite{krizhevsky2009learning}, which maintain a resolution of 32 $\times$ 32, aligning with state-of-the-art benchmarks. Furthermore, we resize the Tiny ImageNet \cite{le2015tiny} datasets to 64 $\times$ 64 for additional experimentation. The supplementary materials provide more detailed dataset information.
\begin{table*}[t]
\centering
\scriptsize
\renewcommand{\arraystretch}{1.1}
\resizebox{1\linewidth}{!}{
\begin{tabular}{c|ccc|ccc|ccc}
\toprule
Dataset   & \multicolumn{3}{c|}{CIFAR-10} & \multicolumn{3}{c|}{CIFAR-100} & \multicolumn{3}{c}{Tiny ImageNet} \\
IPC & 1 & 10 & 50 & 1 & 10 & 50 & 1 & 10 & 50 \\

Ratio \% & 0.02  & 0.2 & 1 & 0.2 & 2  & 10 & 0.2 & 2 & 10  \\
\midrule

Random       
 & $14.4_{\pm2.0}$ & $26.0_{\pm1.2}$ & $43.4±_{\pm1.0}$ & $4.2±_{\pm0.3}$  & $14.6_{\pm0.5}$ & $30.0_{\pm0.4}$ & $1.4_{\pm0.1}$ & $5.0_{\pm0.2}$ & $15.0_{\pm0.4}$\\

Herding \cite{welling2009herding}      
 & $21.5_{\pm1.2}$ & $31.6_{\pm0.7}$ & $40.4_{\pm0.6}$ & $8.3_{\pm0.3}$  & $17.3_{\pm0.3}$ & $33.7_{\pm0.5}$ & $2.8_{\pm0.2}$ & $6.3_{\pm0.2}$ & $16.7_{\pm0.3}$\\

K-Center \cite{sener2018active}    
 & $21.5_{\pm1.3}$ & $14.7_{\pm0.9}$ & $27.0_{\pm1.4}$ & $8.4_{\pm0.3}$  & $17.3_{\pm0.3}$ & $30.5_{\pm0.3}$ & - & -  & - \\

Forgetting \cite{toneva2018empirical}       
 & $13.5_{\pm1.2}$ & $23.3_{\pm1.0}$ & $23.3_{\pm1.1}$ & $4.5_{\pm0.2}$ & $15.1_{\pm0.3}$ & - & $1.6_{\pm0.1}$ & $5.1_{\pm0.2}$ &  $15.0_{\pm0.3}$\\
\hline
{DD}$^\dagger$\cite{wang2018dataset}       
 & - & $36.8_{\pm1.2}$ & - & - & -  & - & - & - & - \\

{LD}$^\dagger$\cite{bohdal2020flexible}           
 & $25.7_{\pm0.7}$ & $38.3_{\pm0.4}$ & $42.5_{\pm0.4}$ & $11.5_{\pm0.4}$  & - & - & - & - \\
 
{DC} \cite{zhao2021dataset}      
 & $28.3_{\pm0.5}$ & $44.9_{\pm0.5}$ & $53.9_{\pm0.5}$ & $12.8_{\pm0.3}$ & $25.2_{\pm0.3}$ & $30.6_{\pm0.6}$ & $5.3_{\pm0.1}$ & $12.9_{\pm0.1}$ & $12.7_{\pm0.4}$\\

{DCC} \cite{lee2022dataset}      
 & $32.9_{\pm0.8}$ & $49.4_{\pm0.5}$ & $61.6_{\pm0.4}$ & $13.3_{\pm0.3}$ & $30.6_{\pm0.4}$ & - & - & - & -\\

{DSA} \cite{zhao2021dataset2}     
 & $28.8_{\pm0.7}$ & $52.1_{\pm0.5}$ & $60.6_{\pm0.5}$ & $13.9_{\pm0.3}$ & $32.3_{\pm0.3}$ & $42.8_{\pm0.4}$ & $5.7_{\pm0.1}$ & $16.3_{\pm0.2}$ & $15.1_{\pm0.2}$\\

{DM} \cite{zhao2023dataset}    
 & $26.0_{\pm0.8}$ & $48.9_{\pm0.6}$ & $63.0_{\pm0.4}$ & $11.4_{\pm0.3}$ & $29.7_{\pm0.3}$ & $43.6_{\pm0.4}$ & $3.9_{\pm0.2}$ & $12.9_{\pm0.4}$ & $25.3_{\pm0.2}$ \\

{GLaD} \cite{cazenavette2023generalizing}        
& $28.0_{\pm0.8}$ & $46.7_{\pm0.5}$ & $59.9_{\pm0.7}$ & - & - & -  & - & - & - \\

CAFE \cite{wang2022cafe}       
& $30.3_{\pm1.1}$ & $46.3_{\pm0.6}$ & $55.5_{\pm0.6}$ & $12.9_{\pm0.3}$ & $27.8_{\pm0.3}$ & $37.9_{\pm0.3}$ & - & - & - \\

CAFE+DSA \cite{wang2022cafe}        
& $31.6_{\pm0.8}$ & $50.9_{\pm0.5}$ & $62.3_{\pm0.4}$ & $14.0_{\pm0.3}$ & $31.5_{\pm0.2}$ & $42.9_{\pm0.2}$  & - & - & - \\

{VIG} \cite{loo2023dataset}        
& $26.5_{\pm1.2}$ & $54.6_{\pm0.1}$ & $35.6_{\pm0.6}$ & $17.8_{\pm0.1}$ & $29.3_{\pm0.1}$ & -  & - & - & - \\

{KIP} \cite{nguyen2021dataset}        
& $29.8_{\pm1.0}$ & $46.1_{\pm0.7}$ & $53.2_{\pm0.7}$ & $12.0_{\pm0.2}$ & $29.0_{\pm0.3}$ & -  & - & - & - \\

{DAM} \cite{sajedi2023datadam}
 & $32.0_{\pm1.2}$ & $54.2_{\pm0.8}$ & $67.0_{\pm0.4}$ & $14.5_{\pm0.5}$ & $34.8_{\pm0.5}$ & $49.4_{\pm0.3}$ & $8.3_{\pm0.4}$ & $18.7_{\pm0.3}$ & $28.7_{\pm0.3}$\\

\textbf{ATOM} (Ours)         
 & $\mathbf{34.8}_{\pm1.0}$ & $\mathbf{57.9}_{\pm0.7}$
& $\mathbf{68.8}_{\pm0.5}$ & $\mathbf{18.1}_{\pm0.4}$
& $\mathbf{35.7}_{\pm0.4}$ & $\mathbf{50.2}_{\pm0.3}$ 
& $\mathbf{9.1}_{\pm0.2}$ & $\mathbf{19.5}_{\pm0.4}$ & $\mathbf{29.1}_{\pm0.3}$ \\  \midrule

% \textbf{D2M} (Ours) 
% & $\mathbf{50.2_{\pm0.3}}$ & $\mathbf{67.8_{\pm0.5}}$ & $\mathbf{74.4_{\pm0.3}}$
% & $\mathbf{29.8_{\pm0.4}}$ & $\mathbf{46.6_{\pm0.3}}$ & $\mathbf{51.2_{\pm0.2}}$
% & $\mathbf{16.7_{\pm0.2}}$ & $\mathbf{26.1_{\pm0.4}}$ & $\mathbf{30.1_{\pm0.2}}$\\ \midrule

Full Dataset & \multicolumn{3}{c|}{$84.8_{\pm0.1}$} & \multicolumn{3}{c|}{$56.2_{\pm0.3}$} & \multicolumn{3}{c}{$37.6_{\pm0.4}$} \\ \bottomrule
\end{tabular}
}
\caption{Comparison with previous dataset distillation methods on CIFAR-10, CIFAR-100 and Tiny ImageNet. The works {DD}$^\dagger$ and {LD}$^\dagger$ use AlexNet \cite{krizhevsky2017imagenet} for CIFAR-10 dataset. All other methods use ConvNet for training and evaluation. \textbf{Bold entries} are the best results.}
\label{tab:cv}
\end{table*}

\textbf{Network Architectures.} We employ a ConvNet architecture \cite{gidaris2018dynamic} for distillation, following prior studies. The default ConvNet comprises three convolutional blocks, each consisting of a 128-kernel 3 $\times$ 3 convolutional layer, instance normalization, ReLU activation, and 3 $\times$ 3 average pooling with a stride of 2. To accommodate the increased resolutions in Tiny ImageNet, we append a fourth convolutional block. Network parameters are initialized using normal initialization \cite{he2015delving} in all experiments.

% We employ a ConvNet architecture \cite{gidaris2018dynamic} for distillation, following prior studies. The default ConvNet comprises three convolutional blocks, with an additional fourth block added to accommodate the higher resolutions in Tiny ImageNet. Network parameters are initialized using normal initialization \cite{he2015delving} in all experiments.

\textbf{Evaluation Protocol.} We evaluate the methods using standard measures from previous studies \cite{zhao2023dataset, zhao2021dataset, wang2022cafe, zhao2021dataset2, sajedi2023datadam}. Five sets of synthetic images are generated from a real training dataset with 1, 10, and 50 images per class. Then, 20 neural network models are trained on each synthetic set using an SGD optimizer with a fixed learning rate of 0.01. Each experiment reports the mean and standard deviation values for 100 models to assess the efficacy of distilled datasets. Furthermore, computational costs are assessed by calculating run-time per step over 100 iterations, as well as peak GPU memory usage during 100 iterations of training.

% We evaluate methods using standard measures from previous studies \cite{zhao2023dataset, zhao2021dataset, wang2022cafe, zhao2021dataset2, sajedi2023datadam}. Five sets of synthetic images are generated from a real training dataset with 1, 10, and 50 images per class. Afterward, 20 neural network models are trained on each synthetic set using an SGD optimizer with a fixed learning rate of 0.01. Each experiment reports mean and standard deviation values for 100 models. Additionally, computational costs are evaluated by measuring runtime per step over 100 iterations and peak GPU memory usage during training.

\textbf{Implementation Details.} We use the SGD optimizer with a fixed learning rate of 1 to learn synthetic datasets containing 1, 10, and 50 IPCs over 8000 iterations with task balances ($\lambda$) set at 0.01. Previous works have shown that $p_s=4$ is sufficient for spatial attention matching \cite{sajedi2023datadam}. As such we set our default case as: $p_c = p_s = 4$. This is further ablated in \Cref{sec:ablation_power}. We adopt differentiable augmentation for both training and evaluating the synthetic set, following \cite{zhao2021dataset, sajedi2023datadam}. For dataset reprocessing, we utilized the Kornia implementation of Zero Component Analysis (ZCA) with default parameters, following previous works \cite{nguyen2021dataset, cazenavette2022dataset, sajedi2023datadam}. All experiments are performed on a single A100 GPU with 80 GB of memory. Further hyperparameter details can be found in the supplementary materials.

\textbf{Competitive Methods.} In this paper, we compare the empirical results of \texttt{ATOM} on three computer vision datasets: CIFAR10/100 and TinyImageNet. We evaluate ATOM against four corset selection approaches and thirteen distillation methods for training set synthesis. The corset selection methods include Random selection \cite{rebuffi2017icarl}, Herding \cite{castro2018end, belouadah2020scail}, K-Center \cite{sener2018active}, and Forgetting \cite{tonevaempirical}. We also compare our approach with state-of-the-art distillation methods, including Dataset Distillation \cite{wang2018dataset} (DD), Flexible Dataset Distillation \cite{bohdal2020flexible} (LD), Dataset Condensation \cite{zhao2021dataset} (DC), Dataset Condensation with Contrastive (DCC) \cite{lee2022dataset}, Dataset Condensation with Differentiable Siamese Augmentation \cite{zhao2021dataset2} (DSA), Distribution Matching \cite{zhao2023dataset} (DM), Deep Generative Priors (GLaD), Aligning Features \cite{wang2022cafe} (CAFE), VIG \cite{loo2023dataset}, Kernel Inducing Points \cite{nguyen2021dataset, nguyen2021dataset2} (KIP) and Attention Matching \cite{sajedi2023datadam} (DAM).

\subsection{Comparison with State-of-the-art Methods}  \label{subsec:comparison}

\textbf{Performance Comparison.} In this section, we present a comparative analysis of our method against coreset and dataset distillation approaches focusing on distribution matching related methods. \texttt{ATOM} consistently outperforms these studies, especially at smaller distillation ratios, as shown in \Cref{tab:cv}. Since the goal of dataset distillation is to generate a more compact synthetic set, we emphasize our significant performance improvements at low IPCs. We achieve almost $4\%$ improvement over the previous attention matching framework \cite{sajedi2023datadam}, DataDAM when evaluated on CIFAR-100 at IPC1. Notably, our performance on CIFAR-100 at IPC50 is 50.2\% -- that is nearly 90\% of the baseline accuracy at a mere 10\% of the original dataset. These examples motivate the development of dataset distillation works as downstream models can achieve relatively competitive performance with their baselines at a fraction of the training costs. Our primary objective in this study is to investigate the impact of channel-wise attention within the feature-matching process. Compared to prior attention-based and feature-based methodologies, our findings underscore the significance of channel-wise attention and the \texttt{ATOM} module, as validated also in the ablation studies in \Cref{sec:ablate_loss_components}.

%In \Cref{tab:cv} we compare our method with distillation and subset selection-based approaches. In particular, our methods achieve superior performance across the test suite, with an emphasis on the smaller distillation ratios. Since the goal of dataset distillation is to obtain a smaller synthetic set, we emphasize our performance improvement at low-IPCs. Most notably, we achieve an almost $4\%$ improvement from the previous attention matching work \cite{} on CIFAR-100 at IPC 1. Our goal in this work wasn't to achieve significantly superior performance in every benchmark, but rather to explore the effect that channel-wise attention had in the feature-matching process. Hence, compared with previous attention-based and feature-matching-based works, our results bolster the importance of channel-wise attention and the ATOM module.

\textbf{Cross-architecture Generalization.} \label{cross}
In this section, we assess the generalization capacity of our refined dataset by training various unseen deep neural networks on it and then evaluating their performance on downstream classification tasks. Following established benchmarks \cite{zhao2021dataset, zhao2023dataset, wang2022cafe, sajedi2023datadam}, we examine classic CNN architectures such as AlexNet \cite{krizhevsky2017imagenet}, VGG-11 \cite{simonyan2014very}, ResNet-18 \cite{he2016deep}, and additionally, a standard Vision Transformer (ViT) \cite{dosovitskiyimage}. Specifically, we utilize synthetic images learned from CIFAR-10 with IPC50 using ConvNet as the reference model and subsequently train the aforementioned networks on the refined dataset to assess their performance on downstream tasks. The results, as depicted in \Cref{tab:crsarc}, indicate that \texttt{ATOM} demonstrates superior generalization across a spectrum of architectures. Notably, it achieves a significant performance boost of over 4\% compared to the prior state-of-the-art on ResNet-18 \cite{he2016deep}. This implies that the channel-wise attention mechanism effectively identifies features not only relevant to ConvNet but also to a wider range of deep neural networks, thereby enhancing the refined dataset with this discerned information.
\begin{table}[htp]
\centering
\scriptsize
\resizebox{1\linewidth}{!}{
\begin{tabular}{cccccc|c}
\toprule
 & ConvNet      & AlexNet      & VGG-11          & ResNet-18  & ViT & Avg. \\ 	\midrule

DC \cite{zhao2021dataset}  & ${53.9_{\pm0.5}}$ & ${28.8_{\pm0.7}}$ & ${38.8_{\pm1.1}}$ & ${20.9_{\pm1.0}}$ & ${30.1_{\pm0.5}}$ & ${34.5_{\pm0.8}}$\\
 
CAFE \cite{wang2022cafe}  & ${62.3_{\pm0.4}}$ & ${43.2_{\pm0.4}}$ & ${48.8_{\pm0.5}}$ & ${43.3_{\pm0.7}}$ & ${22.7_{\pm0.7}}$ & ${44.1_{\pm0.5}}$ \\

DSA \cite{zhao2021dataset2}  & $60.6_{\pm0.5}$ & $53.7_{\pm0.6}$ & $51.4_{\pm1.0}$ & $47.8_{\pm0.9}$ & ${43.3_{\pm0.4}}$ & ${51.4_{\pm0.7}}$ \\

DM \cite{zhao2023dataset}   & $63.0_{\pm0.4}$ & $60.1_{\pm0.5}$ & $57.4_{\pm0.8}$ & $52.9_{\pm0.4}$ & ${45.2_{\pm0.4}}$ & ${55.7_{\pm0.5}}$ \\

KIP \cite{nguyen2021dataset}  & $56.9_{\pm0.4}$ & $53.2_{\pm1.6}$ & $53.2_{\pm0.5}$ & $47.6_{\pm0.8}$ & ${18.3_{\pm0.6}}$ & ${45.8_{\pm0.8}}$ \\

% MTT \cite{cazenavette2022dataset} & $66.2_{\pm0.6}$ & $43.9_{\pm0.9}$  & $48.7_{\pm1.3}$ & $60.0_{\pm0.7}$ & ${47.7_{\pm0.6}}$ & ${53.3_{\pm0.8}}$ \\

DAM \cite{sajedi2023datadam} & $67.0_{\pm0.4}$ & $63.9_{\pm0.9}$ & $64.8_{\pm0.5}$ & $60.2_{\pm0.7}$ & ${48.2_{\pm0.8}}$ & ${60.8_{\pm0.7}}$ \\

\midrule
\textbf{ATOM} (Ours)  & $\mathbf{68.8}_{\pm0.4}$ & $\mathbf{64.1}_{\pm0.7}$ & $\mathbf{66.4}_{\pm0.6}$ & $\mathbf{64.5}_{\pm0.6}$ & $\mathbf{49.5}_{\pm0.7}$ & $\mathbf{62.7}_{\pm0.6}$ \\

\bottomrule
\end{tabular}}
\caption{Cross-architecture testing performance (\%) on CIFAR-10 with 50 images per class. The ConvNet architecture is employed for distillation. \textbf{Bold entries} are the best results.}
\label{tab:crsarc}
\end{table}

% \input{tables/thesis}
% In this section, we compare the generalization of our distilled dataset by training various other deep neural networks on it and measuring downstream testing performance. Consistent with previous state of the art \cite{} we include AlexNet \cite{}, VGG-11 \cite{}, ResNet-18 \cite{} and additionally a standard Vision Transformer (ViT\cite{}). We use the generated images from CIFAR-10 with IPC50 with ConvNet as the matching model and train these subsequent networks on the distilled dataset for measuring downstream performance. In \Cref{tab:crsarc} we show that ATOM has superior generalization performance on a variety of of networks, yielding a significant improvement of $>4\%$ on ResNet-18 when compared to previous state of the art. This indicates that the channel-wise attention was able to further identify features not only relevant to ConvNet, but also to a wider variety of deep neural networks, and distill these features into the learned dataset.

\begin{table}[h]
  \centering
  \setlength{\abovecaptionskip}{0.1cm}
  \resizebox{0.48\textwidth}{!}{
  % \begin{threeparttable}
  \begin{tabular}{c|ccc|ccc}
    \toprule
    \multirow{2}{*}{Method}&\multicolumn{3}{|c}{Run Time (Sec.)}&\multicolumn{3}{|c}{GPU memory (MB)}\\
    & IPC1 & IPC10 & IPC50 & IPC1 & IPC10 & IPC50 \\
    \midrule
    DC \cite{zhao2021dataset}& $0.16_{\pm0.01}$ & $3.31_{\pm0.02}$ & $15.74_{\pm0.10}$ & 3515 & 3621 & 4527\\
    DSA \cite{zhao2021dataset2} & $0.22_{\pm0.02}$ & $4.47_{\pm0.12}$ & $20.13_{\pm0.58}$ & 3513 & 3639 & 4539\\
    DM \cite{zhao2023dataset} & $0.08_{\pm0.02}$ & $0.08_{\pm0.02}$ & $0.08_{\pm0.02}$ & 3323 & 3455 & 3605\\
    MTT \cite{cazenavette2022dataset} & $0.36_{\pm0.23}$& $0.40_{\pm0.20}$ & OOM & 2711 & 8049 & OOM\\
    DAM \cite{sajedi2023datadam} & $0.09_{\pm0.01}$ & $0.08_{\pm0.01}$ & $0.16_{\pm0.04}$ & 3452 & 3561 & 3724\\
    $\textbf{ATOM}^\dagger$ (Ours)& $0.08_{\pm0.02}$ & $0.08_{\pm0.02}$ & $0.13_{\pm0.03}$ & 3152  & 3263  & 4151\\
    \textbf{ATOM} (Ours)& $0.10_{\pm0.02}$ & $0.10_{\pm0.01}$ & $0.17_{\pm0.02}$ & 3601 & 4314 & 5134 \\
    \bottomrule
  \end{tabular}
  }
  \caption{Comparisons of training time and GPU memory usage for prior dataset distillation methods. Run time is averaged per step over 100 iterations, while GPU memory usage is reported as peak memory during the same 100 iterations of training on an A100 GPU for CIFAR-10. Methods that surpass the GPU memory threshold and fail to run are denoted as OOM (out-of-memory). $\textbf{ATOM}^\dagger$ represents our method with on-channel attention, hence offering a better tradeoff in computational complexity.}
    \label{tab:runtime}
\end{table}
\textbf{Distillation Cost Analysis.}
In this section, we delve into an examination of the training costs required for the distillation process. Although the main goal of dataset distillation is to reduce training costs across different applications such as neural architecture search and continual learning, the distillation technique itself must be efficient, enabling smooth operation on consumer-grade hardware. Approaches such as DC, DSA introduce additional computational overhead due to bi-level optimization. Methods such as MTT achieve higher performance but incur the cost of training expert models in addition to overhead in the distillation process (see Table~\ref{tab:runtime}), for equitable comparison, we focus on distribution matching strategies. Within this domain, our method, akin to DM and DAM, capitalizes on randomly initialized networks, obviating the need for training and thereby reducing the computational cost per step involved in the matching stage. As illustrated in \Cref{tab:runtime} utilizing solely the channel-based $\texttt{ATOM}^{\dagger}$ decreases the computational burden of matching compared to the default \texttt{ATOM} configuration. This efficiency is crucial, as channel-wise attention offers a more effective distillation process while maintaining superior performance (refer to \Cref{sec:ablate_loss_components}).

\begin{figure}[t]
\centering
    {\includegraphics[width=0.5\textwidth]{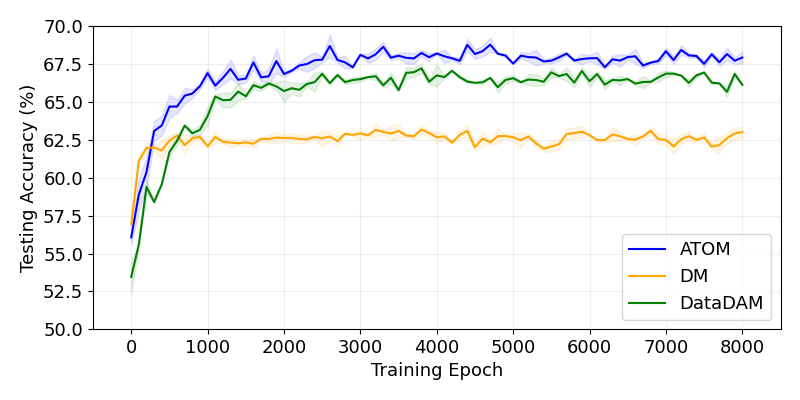}}
    \caption{Test accuracy evolution of synthetic image learning on CIFAR10 with IPC50 for ATOM (ours), DM \cite{zhao2023dataset} and DataDAM \cite{sajedi2023datadam}.}
    \label{fig:convergence}
\end{figure}

\textbf{Convergence Speed Analysis.}
In ~\Cref{fig:convergence}, we plot the downstream testing accuracy evolution for the synthetic images on CIFAR10 IPC50. Comparing with previous methods, DM \cite{zhao2023dataset} and DataDAM \cite{sajedi2023datadam}, we can explicitly see an improvement in convergence speed and a significantly higher steady state achieved with the ATOM framework. Our included convergence analysis supports the practicality of our method and the consistency to which we outperform previous baselines.

\subsection{Ablation Studies and Analysis} \label{ablation}

\begin{figure*}[t]
\centering
    {\includegraphics[width=1\textwidth]{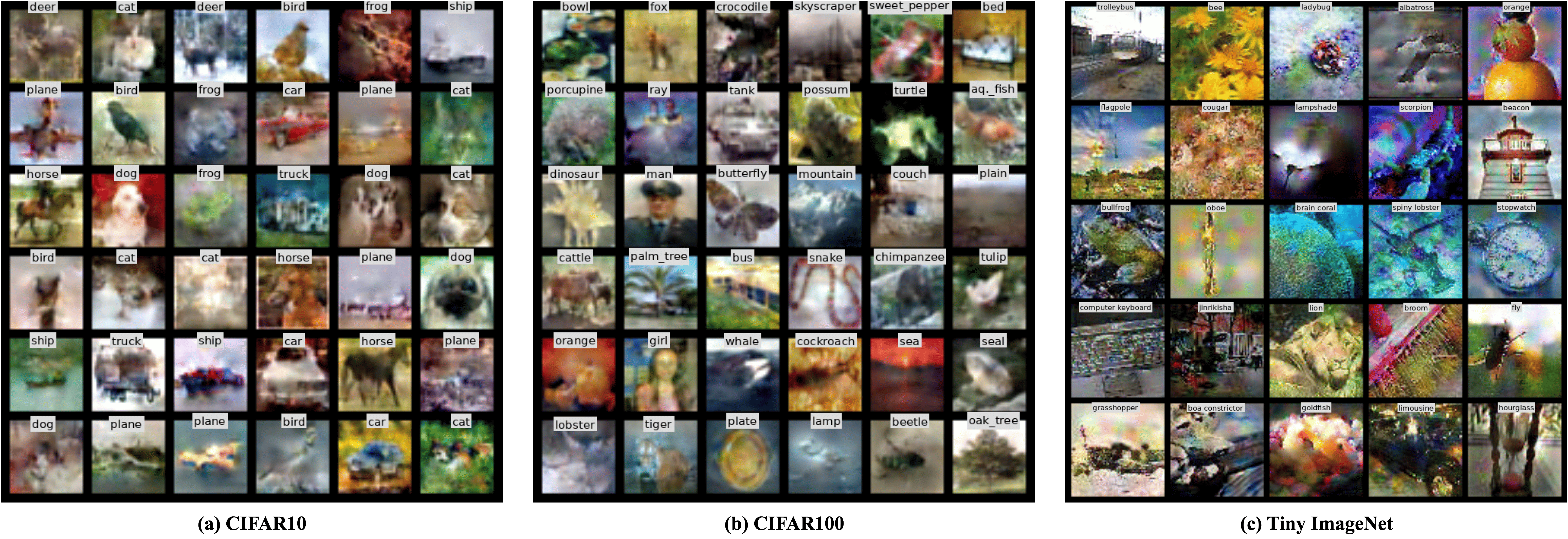}}
    \caption{Sample learned synthetic images for CIFAR-10/100 (32$\times$32 resolution) IPC10 and TinyImageNet (64$\times$64 resolution) IPC 1.}
    \label{fig:atom}
\end{figure*}

\textbf{Evaluation of loss components in ATOM.}
\label{sec:ablate_loss_components}
In \Cref{tab:component}, we evaluate the effect of different attention-matching mechanisms with respect to pure feature matching in intermediate layers and distribution matching in the final layer ($\mathcal{L}_{\text{MMD}}$). The results clearly demonstrate that attention-matching improves the performance of the distillation process. In particular, the attention-matching process improves feature matching by $8.0\%$. Further, it seems that channel attention is able to capture the majority of relevant information from the intermediate features, as evidenced by an improvement of over $1.5\%$ from spatial attention matching. Ultimately, this provides an incentive to favor channel attention in the distillation process.
% \textcolor{red}{Hence, this study provides insight into the type of information that can be captured from attention and rather than channel-wise attention can capture the majority of the information content from the intermediate features.}

% In \Cref{tab:component}, we evaluate the effect of each attention matching with respect to the pure feature/distribution matching in the final layer. It is quite evident that attention matching improves the performance of the distillation process. However, it also seems that the majority of the improved performance can indeed be attributed to channel-wise attention matching.  

% \textbf{Evaluation of loss components in ATOM.}
\begin{table}[h]
% \vspace{-13pt}
\centering
\setlength{\abovecaptionskip}{0.1cm}
\renewcommand\arraystretch{0.9}
\scriptsize
        \setlength{\tabcolsep}{8pt}
% \setlength{\abovecaptionskip}{0.cm}
% \resizebox{\linewidth}{0.1\linewidth}{%
\resizebox{1\linewidth}{!}{
\begin{tabular}{cccc|c}
	\toprule
	$\mathcal{L}_{\text{MMD}}$ &  Feature Map & $\text{Spatial Atn.}$ & $\text{Channel Atn.}$ &
 Performance (\%)    \\
	\midrule
  
	\checkmark & - & - & - & $63.0_{\pm0.4}$ \\
             & \checkmark & - & - & $60.8_{\pm0.6}$ \\
  \checkmark & - & \checkmark & - & $67.0_{\pm0.7}$   \\
  \checkmark & - & - &\checkmark & $68.6_{\pm0.3}$  \\
  
	 \checkmark & - & \checkmark &  \checkmark & $\mathbf{68.8}_{\pm0.5}$ \\
      % \checkmark & \checkmark & \checkmark &   &  *\\
      % \checkmark & \checkmark & \checkmark & \checkmark  &  *\\
% 	\checkmark &            &            & 50.4 & \bf 50.9 \\
% 	           & \checkmark &            & 60.6 & \bf 61.4 \\
% 	           &            & \checkmark & 44.9 & \bf 45.2 \\
% 	           &            &            & 45.5 & \bf 45.9 \\
\bottomrule
\end{tabular}}
% }
\caption{Evaluation of loss components and attention components in \texttt{ATOM} using CIFAR-10 with IPC50.}
\label{tab:component}
\end{table}

\textbf{Evaluating attention balance in ATOM.}
\label{sec:ablation_power}
In this section, we evaluate the balance between spatial and channel-wise attention through the power value $p$. Referencing ~\Cref{eq:spatial} and ~\Cref{eq:channel}, modulating the values of $p_{s}$ and $p_{c}$ ultimately affects the balance of spatial and channel-wise attention in $\mathcal{L}_{\text{ATOM}}$. In Table \ref{tab:power}, we examine the impact of different exponentiation powers $p$ in the attention-matching mechanisms. Specifically, we conduct a grid-based search to investigate how varying the exponentiation of spatial ($p_{s}$) and channel ($p_{c}$) attention influences subsequent performance. Our findings reveal that optimal performance (nearly $1\%$ improvement over our default) occurs when the exponentiation for channel attention significantly exceeds that of spatial attention. This suggests that assigning a higher exponential value places greater emphasis on channel-attention matching over spatial-wise matching. This aligns with our observations from the loss component ablation, where channel-wise matching was found to encapsulate the majority of information within the feature map. Consequently, we deduce that prioritizing channel-wise matching will enhance downstream performance outcomes.

% In \Cref{tab:power}, we evaluate the effect of different powers $p$ (exponentiation) in the attention-matching mechanisms. In particular, we perform a grid-based search to explore how the contrasting exponentiation of spatial and channel attention can affect downstream performance. We find that the performance is best (almost $1\%$ better than our default) when the channel attention exponentiation is significantly larger than that of the spatial attention. Intuitively, a larger exponential would in turn place a greater weight channel-attention matching as opposed to spatial-wise. This concurs with our results in the loss component ablation, where channel-wise matching was deemed to contain the majority of information from the feature map. Hence, we concluded that a greater emphasis on channel-wise matching will positively influence the obtained downstream performance. 

\begin{table}[h]
% \vspace{-13pt}
\centering
\setlength{\abovecaptionskip}{0.1cm}
\renewcommand\arraystretch{1}
\scriptsize
        \setlength{\tabcolsep}{8pt}
\begin{tabular}{ccccc}
	\toprule
     \multirow{2}{*}{Channel Attention $p_{c}$} & \multicolumn{4}{c}{Spatial Attention $p_{s}$}\\
    % \multirow{2}{*}{8}
	 & 1 & 2 & 4 & 8\\
    % $&  & &  & \\
	\midrule
  
	1 & 57.4\% & 57.5\% &57.0\% & 56.2\%\\
    2 & 58.2\% & 57.5\% &57.2\% & 56.3\%\\
    4 & 58.4\% & 58.5\% &57.9\% & 57.6\%\\
	8 & \bf{58.8\%} & 58.7\% &58.2\% & 57.8\%\\
\bottomrule
\end{tabular}
% }
\caption{Evaluation of power values in the spatial and channel attention computations for $\mathcal{L}_\text{ATOM}$ using CIFAR-10 with IPC10.}
\label{tab:power}
\end{table}

\textbf{Visualization of Synthetic Images.} We include samples of our distilled images in \Cref{fig:atom}. The images appear to be interleaved with artifacts that assimilate the background and object information into a mixed collage-like appearance. The synthetic images effectively capture the correlation between background and object elements, suggesting their potential for generalizability across various architectures, as empirically verified in \Cref{tab:crsarc}. Additional visualizations are available in the supplementary material.
%We include samples of our distilled images in \Cref{fig:DataDAM}. The images appear to be interleaved with artifacts that assimilate the background and object information into a mixed collage-like appearance. The synthetic images appear to capture the said correlation between the background and objects that appear to be data-centric hence intuitively making the synthetic data generalizable to a variety of architectures. This was confirmed empirically in \Cref{tab:crsarc}. We include more visualizations in the supplementary material.

\subsection{Applications} \label{app}

\textbf{Neural Architecture Search.} In Table \Cref{tab:nas} we leverage our distilled synthetic datasets as proxy sets to accelerate Neural Architecture Search. In line with previous state-of-the-art, \cite{sajedi2023datadam, zhao2021dataset, zhao2021dataset2}, we outline our architectural search space, comprising 720 ConvNets on the CIFAR-10 dataset. We commence with a foundational ConvNet and devise a consistent grid, varying in depth $D \in$ \{1, 2, 3, 4\}, width $W \in$ \{32, 64, 128, 256\}, activation function $A \in$ \{\text{Sigmoid, ReLU, LeakyReLU}\}, normalization technique $N \in$ \{None, BatchNorm, LayerNorm, InstanceNorm, GroupNorm\}, and pooling operation $P \in$ \{None, MaxPooling, AvgPooling\}. Additionally, we benchmark our approach against several state-of-the-art methods, including Random, DSA \cite{zhao2021dataset}, DM \cite{zhao2023dataset}, CAFE \cite{wang2022cafe}, DAM \cite{sajedi2023datadam}, and Early-Stopping. Our method demonstrates superior performance, accompanied by a heightened Spearman's correlation (0.75), thereby reinforcing the robustness of \texttt{ATOM} and its potential in neural architecture search.
% In Table \Cref{tab:nas} we leverage our distilled synthetic datasets as proxy sets to accelerate Neural Architecture Search. Following previous state-of-the-art \cite{}, we describe our architectural search space (network depth, width, activations, normalization, and pooling) in the Supplementary Material. Moreover, we compare the performance of our method with several state-of-the-art benchmarks including Random, DSA, DM, CAFE, and Early-Stopping. We show a superior performance of **\% with a higher Spearman's correlation thus bolstering the robustness of our method and its applications in neural architecture search. 

\begin{table}[h]
\centering
\scriptsize
\setlength{\abovecaptionskip}{0.1cm}
\resizebox{1\linewidth}{!}{
\begin{tabular}{cccccccc|c}
\toprule
                    & Random            & DSA               & DM             & CAFE & DAM & \textbf{ATOM} & Early-stopping & Full Dataset \\ \midrule
Performance (\%)    &        88.9         &             87.2   &         87.2   &  83.6 &  \bf{89.0}  & 88.9 &   88.9 & 89.2  \\
Correlation    &        0.70         &        0.66        &       0.71     &  0.59 &  0.72   &  \bf{0.75}  &  0.69  &  1.00\\
Time cost (min)     & 206.4  &   206.4    & 206.6 &     206.4  &   206.4  &  206.4  &   206.2    & 5168.9 \\ 
Storage (imgs)      & \bf{500}          & \bf{500}          & \bf{500}        &   \bf{500}  &   \bf{500}  &  \bf{500} & $5\times 10^4$    & $5\times 10^4$   \\ \bottomrule
\end{tabular}}
\caption{{Neural architecture search on CIFAR-10 with IPC50.}}
\label{tab:nas}
\end{table}

\section{Limitations}
Many studies in dataset distillation encounter a constraint known as \textit{re-distillation costs} \cite{wang2023dim, he2024you, he2024multisize}. This limitation becomes apparent when adjusting the number of images per class (IPC) or the distillation ratios. Like most other distillation methods, our approach requires re-distillation on the updated setting configuration, which limits flexibility regarding configuration changes and storage allocation. Additionally, we observed in \Cref{tab:crsarc} that dataset distillation methods often struggle with generalizing to transformer architectures. Despite \texttt{ATOM} outperforming other methods, there is still a noticeable performance drop compared to convolutional neural networks. This suggests that the effectiveness of transformers for downstream training might be constrained by the distilled data.

% Many works in the field of dataset distillation are limited by the concept of re-distillation costs \cite{}. When varying the IPC or dataset, our method, as with most other distillation methods, would need to be re-run on the new configuration. Unfortunately, this is a limitation of many distillation works, making them less flexible to configuration/storage allocation changes. Similarly, we leave this as an open problem in the field. Further, we note that in \Cref{tab:crsarc}, distillation methods appear to exhibit poor generalization to transformer architectures. Despite ATOM outperforming the other methods, we still notice a performance drop from convolutional neural networks, hence the performance of transformers for downstream training may be limited by the distilled data.

\section{Conclusion}
In this work, we introduced an Attention Mixer (\texttt{ATOM}) for efficient dataset distillation. Previous approaches have struggled with marginal performance gains, obfuscating channel-wise information, and high computational overheads. \texttt{ATOM} addresses these issues by effectively combining information from different attention mechanisms, facilitating a more informative distillation process with untrained neural networks. Our approach utilizes a broader receptive field to capture spatial information while preserving distinct content information at the channel level, thus better aligning synthetic and real datasets. By capturing information across intermediate layers, \texttt{ATOM} facilitates multi-scale distillation. We demonstrated the superior performance of \texttt{ATOM} on standard distillation benchmarks and its favorable performance across multiple architectures. We conducted several ablative studies to justify the design choices behind \texttt{ATOM}. Furthermore, we applied our distilled data to Neural Architecture Search, showing a superior correlation with the real large-scale dataset. In the future, we aim to extend attention mixing to various downstream tasks, including image segmentation and localizations. We also hope to address limitations of \texttt{ATOM}, such as re-distillation costs and cross-architecture generalizations on transformers.
% In this work, we introduced an Attention Mixer (ATOM) for efficient dataset distillation. Motivated by previous works limitations in marginal performance improvements\cite{}, obfuscating channel-wise information \cite{}, and heavy computational costs \cite{}, ATOM was presented to effectively leverage joint information from different attention schemes for a more informative distillation process with untrained neural networks. Our attention mixer leverages a wider receptive field by capturing information spatially, paired with distinct content information obtained channel-wise, to best match our synthetic and real datasets. Capturing information in the intermediate layers enables the distillation of multi-scale information. ATOM has shown superior performance on the common distillation test suite and performs favorably across multiple architectures. Several ablative studies are performed to justify design choices in creating ATOM. Finally, we include the results of applying our distilled data to Neural Architecture Search, wherein our synthetic data offers superior correlation. In the future, we hope to extend the concepts of attention mixing to various dense-prediction tasks.

{
    \small
    \bibliographystyle{ieeenat_fullname}
    \bibliography{main}
}

 \clearpage
\setcounter{page}{1}
\maketitlesupplementary

\section{Implementation Details}
\label{sec:rationale}

\subsection{Datasets}
We conducted experiments on three main datasets: CIFAR10/100 \cite{krizhevsky2009learning} and TinyImageNet \cite{le2015tiny}. These datasets are considered single-label multi-class; hence, each image has exactly one class label. The CIFAR10/100 are conventional computer vision benchmarking datasets comprising 32$\times$32 colored natural images. They consist of 10 coarse-grained labels (CIFAR10) and 100 fine-grained labels (CIFAR100), each with 50,000 training samples and 10,000 test samples. The CIFAR10 classes include "Airplane", "Car", "Bird", "Cat", "Deer", "Dog", "Frog", "Horse", "Ship", and "Truck". The TinyImageNet dataset, a subset of ImageNet-1K \cite{deng2009imagenet} with 200 classes, contains 100,000 high-resolution training images and 10,000 test images resized to 64$\times$64. The experiments on these datasets make up the benchmarking for many previous dataset distillation works \cite{zhao2021dataset, wang2022cafe, sajedi2023datadam, cazenavette2022dataset, cazenavette2023generalizing, zhou2023dataset}.

\subsection{Dataset Pre-processing}
We applied the standardized preprocessing techniques to all datasets, following the guidelines provided in DM \cite{zhao2023dataset} and DataDAM \cite{sajedi2023datadam}. Following previous works, we apply the default Differentiable Siamese Augmentation (DSA) \cite{zhao2021dataset2} scheme during distillation and evaluation. Specifically for the CIFAR10/100 datasets, we integrated Kornia zero-phase component analysis (ZCA) whitening, following the parameters outlined in \cite{cazenavette2022dataset, sajedi2023datadam}. Similar to DataDAM \cite{sajedi2023datadam}, we opted against ZCA for TinyImagenet due to the computational bottlenecks associated with full-scale ZCA transformation on a larger dataset with double the resolution. Note that we visualized the distilled images by directly applying the inverse transformation based on the corresponding data pre-processing, without any additional modifications.

\subsection{Hyperparameters}
Our method conveniently introduces only one additional hyperparameter: the power term in channel attention, \ie $p_c$. All the other hyperparameters used in our method are directly inherited from the published work, DataDAM \cite{sajedi2023datadam}. Therefore, we include an updated hyperparameter table in \Cref{tab: hyperparameters} aggregating our power term with the remaining pre-set hyperparameters. In the main paper, we discussed the effect of power terms on both channel- and spatial-wise attention and ultimately found that higher channel attention paired with lower spatial attention works best. However, our default, as stated in the main draft, is $p_c = p_s = 4$. Regarding the distillation and train-val settings, we use the SGD optimizer with a learning rate of 1.0 for learning the synthetic images and a learning rate of 0.01 for training neural network models (for downstream evaluation). For CIFAR10/100 (low-resolution), we use a 3-layer ConvNet; meanwhile, for TinyImagenet (medium-resolution), we use a 4-layer ConvNet, following previous works in the field \cite{zhao2023dataset, sajedi2023datadam, cazenavette2022dataset}. Our batch size for learning the synthetic images was set to 128 due to the computational overhead of a larger matching set.

\subsection{Neural Architecture Search Details}

Following previous works \cite{sajedi2023datadam, zhao2021dataset, zhao2021dataset2, zhao2023dataset}, we define a search space consisting of 720 ConvNets on the CIFAR10 dataset. Models are evaluated on CIFAR10 using our IPC 50 distilled set as a proxy under the neural architecture search (NAS) framework. The architecture search space is constructed as a uniform grid that varies in depth $D \in$ \{1, 2, 3, 4\}, width $W \in$ \{32, 64, 128, 256\}, activation function $A \in$ \{\text{Sigmoid, ReLu, LeakyReLu}\}, normalization technique $N \in$ \{None, BatchNorm, LayerNorm, InstanceNorm, GroupNorm\}, and pooling operation $P \in$ \{None, MaxPooling, AvgPooling\} to create varying versions of the standard ConvNet. These candidate architectures are then evaluated based on their validation performance and ranked accordingly. In the main paper, \Cref{tab:nas} measures various costs and performance metrics associated with each distillation method. Overall distillation improves the computational cost; however, \texttt{ATOM} achieves the highest correlation, which is by far the most ``important`` metric in this NAS search, as it indicates that our proxy set best estimates the original dataset.

\section{Additional Visualizations.} 
\begin{figure*}
    \centering
    \includegraphics[width=\textwidth]{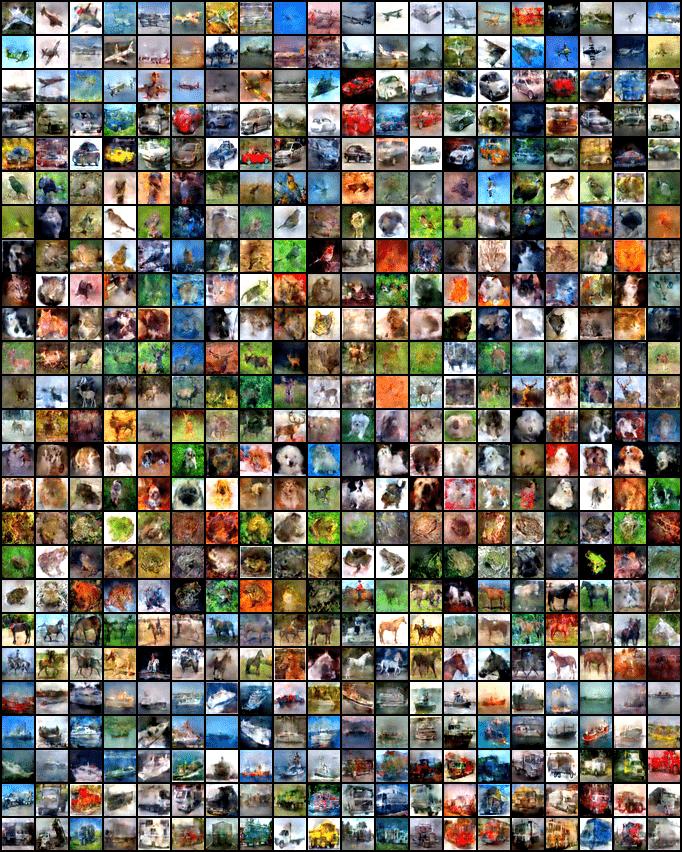}
    \caption{Distilled Image Visualization: CIFAR-10 dataset with IPC 50.}
    \label{fig:cifar10ipc50}
\end{figure*}

\begin{figure*}
    \centering
    \includegraphics[width=\textwidth]{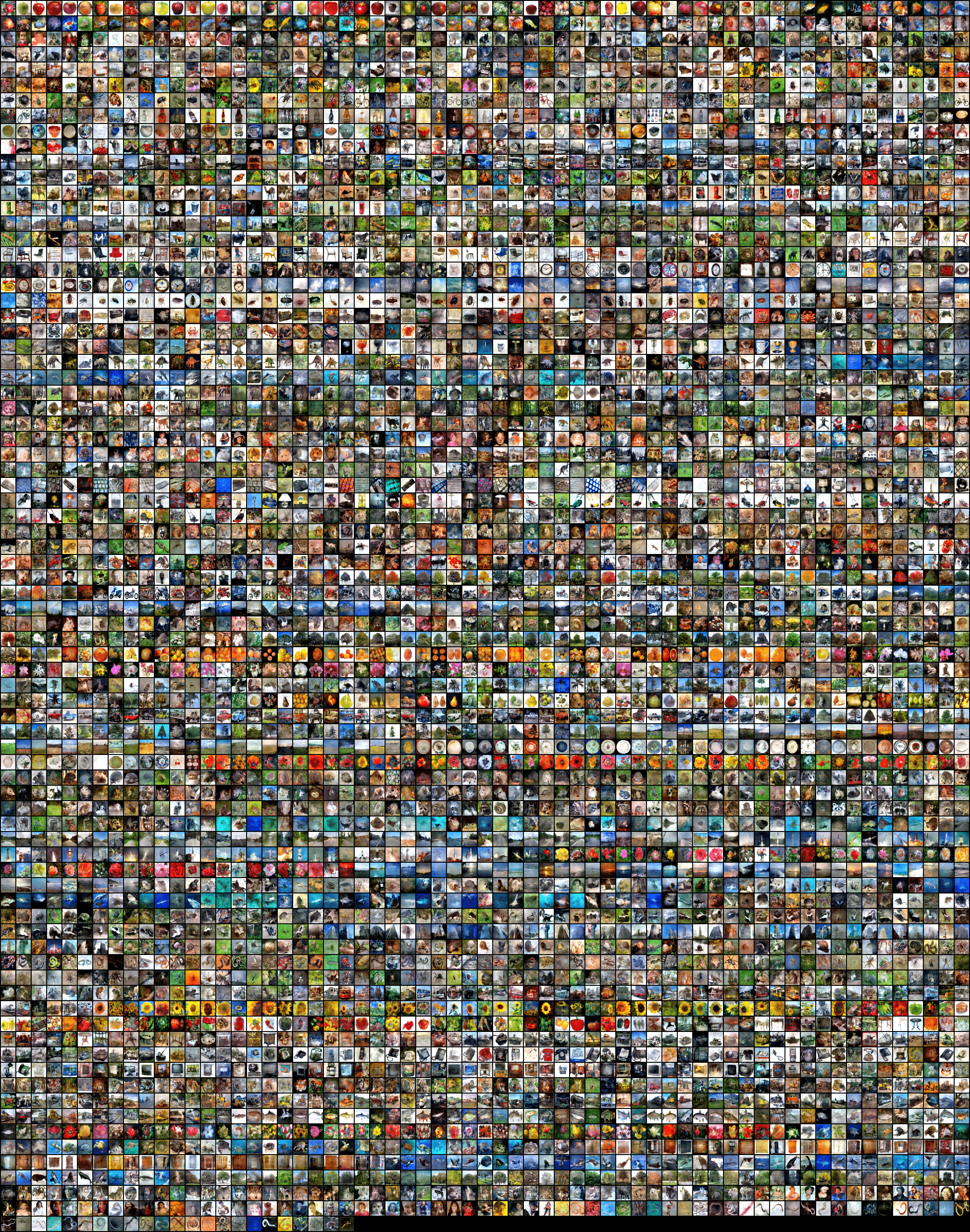}
    \caption{Distilled Image Visualization: CIFAR-100 dataset with IPC 50.}
    \label{fig:cifar100ipc50}
\end{figure*}

\begin{figure*}
    \centering
    \includegraphics[width=\textwidth]{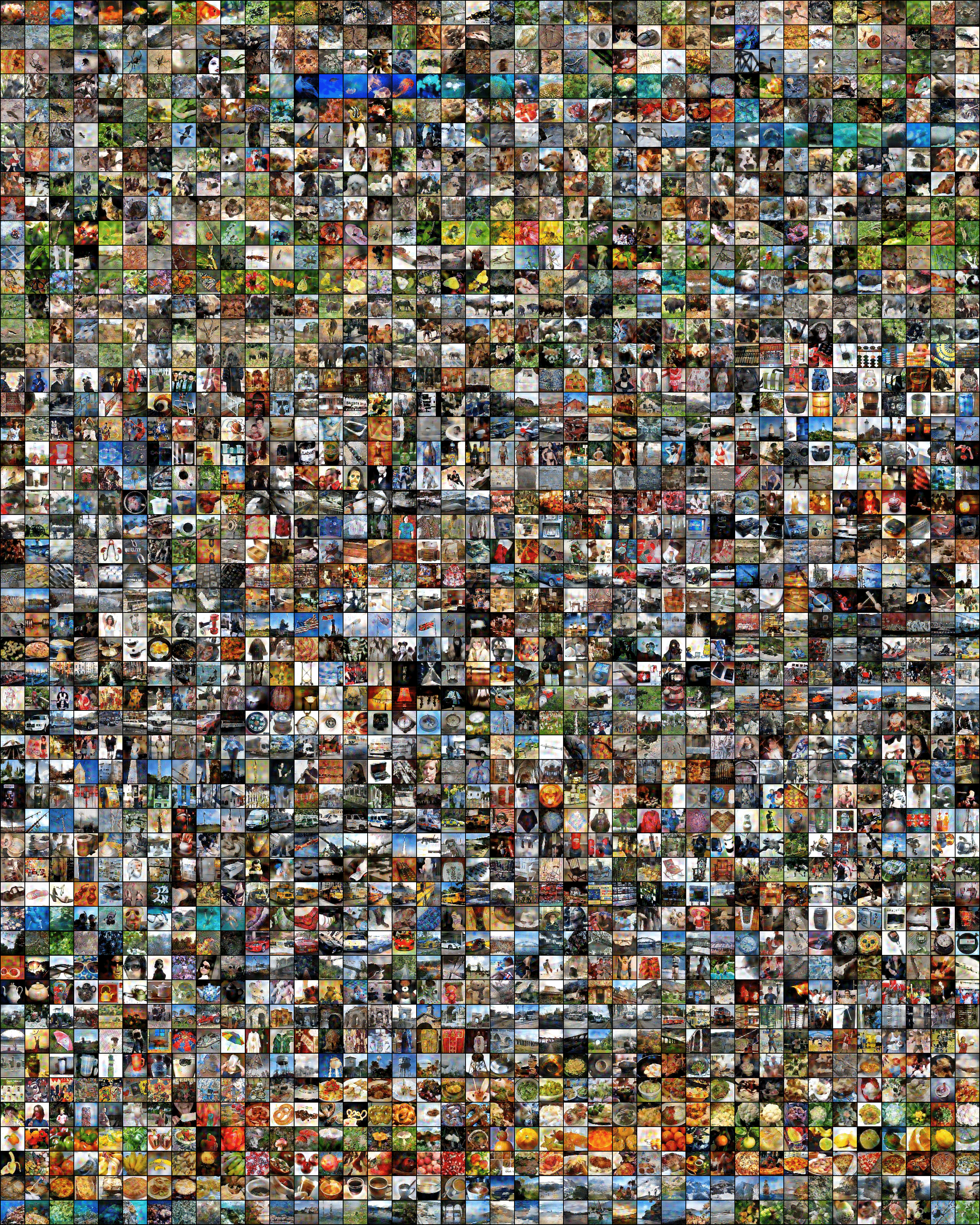}
    \caption{Distilled Image Visualization: TinyImageNet dataset with IPC 10.}
    \label{fig:tinyipc10}
\end{figure*}

We include additional visualizations of our synthetic datasets in \Cref{fig:cifar10ipc50}, \Cref{fig:cifar100ipc50}, \Cref{fig:tinyipc10}. The first two represent CIFAR10/100 at IPC 50, while the third depicts TinyImageNet at IPC 10. Our images highly exhibit learned artifacts from the distillation process that are, in turn, helpful during downstream classification tasks.
%In \Cref{fig:cifar10ipc50}, \Cref{fig:cifar100ipc50}, \Cref{fig:tinyipc10}, we include 3 visuals of our synthetic datasets. The former 2 are CIFAR10/100 at IPC 50, meanwhile the latter is TinyImageNet at IPC 10. Our images highly exhibit learned artifacts from the distillation process that are in turn helpful during downstream training and evaluation.

\begin{table*} [t]
    \centering
    \resizebox{\textwidth}{!}{
    \normalsize{
    \begin{tabular}{c|c|c||c||c}
    \hline  
    \multicolumn{3}{c||}{\textbf{Hyperparameters}} &
    \multirow{2}{*}{} \textbf{Options/} &
    \multirow{2}{*}{\textbf{Value}} \\
    \cline{1-3}
    \textbf{Category} & \textbf{Parameter Name} & \textbf{Description} & \textbf{Range} & \\
    \hline 
    \hline
    
    \multirow{12}{*}{\textbf{Optimization}}
    & \textbf{Learning Rate $\boldsymbol{\eta_{\mathcal{S}}}$ (images) } & Step size towards global/local minima & $(0, 10.0]$ & IPC $\leq$ 50: $ 1.0$ \\ 
    & & &  & IPC $>$ 50: $10.0$\\
    \cline{2-5}

    & \textbf{Learning Rate $\boldsymbol{\eta_{\mathcal{\boldsymbol{\theta}}}}$ (network)} & Step size towards global/local minima & $(0, 1.0]$ & $0.01$ \\ 
    \cline{2-5}
    
    \multirow{4}{*}{} & 
    
    \multirow{2}{*}{\textbf{Optimizer (images)}} & \multirow{2}{*}{Updates synthetic set to approach global/local minima} & SGD with & Momentum: $0.5$ \\
    & & & Momentum & Weight Decay: $0.0$\\
    \cline{2-5}
    & \multirow{2}{*}{\textbf{Optimizer (network)}} & \multirow{2}{*}{Updates model to approach global/local minima} & SGD with & Momentum: $0.9$ \\
    & & & Momentum & Weight Decay: $5e-4$\\
    
    \cline{2-5}
    
    \multirow{3}{*}{} & 

    \multirow{1}{*}{\textbf{Scheduler (images)}} & - & - & - \\
    \cline{2-5}
    &\multirow{2}{*}{\textbf{Scheduler (network)}} & \multirow{2}{*}{Decays the learning rate over epochs} & \multirow{2}{*}{StepLR} & Decay rate: $0.5$ \\
    &&&& Step size: $15.0$\\
    \cline{2-5}
     & \textbf{Iteration Count} & Number of iterations for learning synthetic data & $[1, \infty)$ & 8000\\
    \hline
    \multirow{6}{*}{\textbf{Loss Function}}
    & \multirow{2}{*}{\textbf{Task Balance $\lambda$}} & \multirow{2}{*}{Regularization Multiplier} & \multirow{2}{*}{$[0, \infty)$} & Low Resolution: $0.01$ \\
    &&&& High Resolution: $0.02$\\
    \cline{2-5}
    &  \textbf{Spatial Power Value} \bf{$p_s$} & Exponential power for amplification of spatial attention & $[1, \infty)$ & 4 \\
    \cline{2-5}
    &  \textbf{Channel Power Value} \bf{$p_c$} & Exponential power for amplification of channel attention & $[1, \infty)$ & 4 \\
    \cline{2-5}
    & \textbf{Loss Configuration} & Type of error function used to measure distribution discrepancy & - & Mean Squared Error \\
    \cline{2-5}
    & \textbf{Normalization Type} & Type of normalization used in the SAM module on attention maps & - & L2 \\
    \cline{2-5}
    
    % & \textbf{eta_2}$\mathbb{(\eta_2)}$ & Lagrange Multiplier for MSE & $[0.1, 0.9]$ & $0.7$ \\
    % \cline{2-5}
    
    % & \textbf{skew}$\mathbb{(\alpha)}$ & Measure of symmetry in GJSD & $(0, 1.0]$ & $0.5$ \\
    \hline
    
    % \multirow{5}{*}{\textbf{Color Augmentation}}
    % & \multirow{4}{*}{\textbf{Method}} & \multirow{4}{*}{Method to alter the intensities of the color channels} & YCbCr & \multirow{4}{*}{YCbCr}\\
    % &&& HSV & \\
    % &&& Color Distortion & \\
    % &&& RGB-Jitter &\\
    % \cline{2-5}
    
    % & \textbf{Distortion factor} & The extent of perturbations of the color intensities & (0, 1.0] & 0.3 \\
    % \hline
    % Need to confirm these values
    % Grab Probabilities for each from dict.
    \multirow{8}{*}{\textbf{DSA Augmentations}}
    & \multirow{3}{*}{\textbf{Color}} &  \multirow{3}{*}{Randomly adjust (jitter) the color components of an image} & brightness & 1.0\\
    & & & saturation & 2.0\\
    & & & contrast & 0.5\\
    \cline{2-5}
    
     & \textbf{Crop} & Crops an image with padding & ratio crop pad & 0.125 \\
     \cline{2-5}
     & \textbf{Cutout} & Randomly covers input with a square & cutout ratio & 0.5 \\
    \cline{2-5}
    & \textbf{Flip} & Flips an image with probability p in range: & $(0, 1.0]$ & $0.5$ \\
    \cline{2-5}
    & \textbf{Scale} & Shifts pixels either column-wise or row-wise & scaling ratio & $1.2$ \\
    \cline{2-5}
    & \textbf{Rotate} & Rotates image by certain angle & $0^{\circ} - 360^{\circ}$ & $[-15^{\circ}, +15^{\circ}]$ \\
    \cline{2-5}

    \hline
    
    % \textbf{Dataset} & \textbf{Labels Hierarchy} & Three levels of hierarchy in annotations \cite{ATLAS8953780} & L1, L2, L3Only, L3 & L3Only \\
    % \hline
    
    % \multirow{6}{*}{}
    % & \textbf{Mean layer weights} & \multirow{3}{*}{The weights of linear transformation layer} & \multirow{3}{*}{$\mathbb{R}$} & Random values\\
    % & \textbf{Variance layer weights} &&& from Normal\\
    % \textbf{GMM Layer parameters} & \textbf{Pi layer weights} &&& distribution\\
    % \cline{2-5}
    
    % \textbf{Initialization} & \textbf{Mean layer biases} & \multirow{3}{*}{The biases of linear transformation layer}  &\multirow{3}{*}{Constant}& \multirow{3}{*}{0} \\
    % & \textbf{Variance layer biases} &&&\\
    % & \textbf{Pi layer biases} &&&\\
    % \hline
    
    \multirow{3}{*}{\textbf{Encoder Parameters}} & \textbf{Conv Layer Weights} & The weights of convolutional layers & $\mathbb{R}$ bounded by kernel size & Uniform Distribution  \\
    \cline{2-5}
    
    & \textbf{Activation Function} & The non-linear function at the end of each layer & - & ReLU \\
    \cline{2-5}
    & \textbf{Normalization Layer} & Type of normalization layer used after convolutional blocks & - & InstanceNorm\\
    \hline
    \end{tabular}
    }       }
    \caption{Hyperparameters Details -- boilerplate obtained from DataDAM \cite{sajedi2023datadam}.}
    \label{tab: hyperparameters}
\end{table*}
% % 
% Having the supplementary compiled together with the main paper means that:
% % 
% \begin{itemize}
% \item The supplementary can back-reference sections of the main paper, for example, we can refer to \cref{sec:intro};
% \item The main paper can forward reference sub-sections within the supplementary explicitly (e.g. referring to a particular experiment); 
% \item When submitted to arXiv, the supplementary will already included at the end of the paper.
% \end{itemize}
% % 
% To split the supplementary pages from the main paper, you can use \href{https://support.apple.com/en-ca/guide/preview/prvw11793/mac#:~:text=Delete%20a%20page%20from%20a,or%20choose%20Edit%20%3E%20Delete).}{Preview (on macOS)}, \href{https://www.adobe.com/acrobat/how-to/delete-pages-from-pdf.html#:~:text=Choose%20%E2%80%9CTools%E2%80%9D%20%3E%20%E2%80%9COrganize,or%20pages%20from%20the%20file.}{Adobe Acrobat} (on all OSs), as well as \href{https://superuser.com/questions/517986/is-it-possible-to-delete-some-pages-of-a-pdf-document}{command line tools}.

\end{document}